\documentclass[a4paper]{article}

\usepackage{INTERSPEECH2015}

\usepackage{graphicx}
\usepackage{amssymb,amsmath,bm}
\usepackage{textcomp}

\ninept

\title{Applying GPGPU to Recurrent Neural Network Language Model based Fast Network Search in the Real-Time LVCSR}

\makeatletter
\def\name#1{\gdef\@name{#1\\}}
\makeatother \name{{\em Kyungmin Lee, Chiyoun Park, Ilhwan Kim, Namhoon Kim, and Jaewon Lee}}

\address{DMC R\&D Center, Samsung Electronics, Suwon, Korea \\
  {\small \tt \{karaf.lee, chiyoun.park, ilhwans.kim, namhoon.kim, jwonlee\}@samsung.com}
}

\begin{document}

  \maketitle
  \begin{abstract}
	Recurrent Neural Network Language Models (RNNLMs) have started to be used in various fields of speech recognition due to their outstanding performance.	
	However, the high computational complexity of RNNLMs has been a hurdle in applying the RNNLM to a real-time Large Vocabulary Continuous Speech Recognition (LVCSR).
	In order to accelerate the speed of RNNLM-based network searches during decoding, we apply the General Purpose Graphic Processing Units (GPGPUs).
	This paper proposes a novel method of applying GPGPUs to RNNLM-based graph traversals.
	We have achieved our goal by reducing redundant computations on CPUs and amount of transfer between GPGPUs and CPUs.
	The proposed approach was evaluated on both WSJ corpus and in-house data.
	Experiments shows that the proposed approach achieves the real-time speed in various circumstances while maintaining the Word Error Rate (WER) to be relatively 10\% lower than that of n-gram models.
	
  \end{abstract}
  \noindent{\bf Index Terms}: Recurrent Neural Network, Language Model, General Purpose Graphics Processing Units, Large Vocabulary Continuous Speech Recognition

\section{Introduction}

%RNNLM 장점 및 특징
Recently, the Recurrent Neural Network Language Model (RNNLM) has gained its popularity in the field of Automatic Speech Recognition (ASR). 
%G The 삭제, research는 단수(아마도?)
Various academic research has reported the effectiveness of RNNLMs, which can train unseen contexts by sharing the statistics between words 
%G of which contexts -> whose contexts are / the contexts of which are
with syntactically and semantically similar contexts~\cite{rnnlm0, rnnlm2wfst, rnnlm2n-gram, onepass1, onepass2}. However, heavy computational load of RNNLM over traditional n-gram based approaches 
%G keep it from ... it이 의미하는 바가 없음..
%G an every area -> every area
has been a hurdle in applying the RNNLM to diverse areas of ASR applications. 
%G if -> when,  to be run -> to run.. 
%G 여전히 when/if 구문이 있는 것 자체가 어색..
Especially, when ASR systems are required to run under real-time constraint (i.e., less than 1xRT), 
%G hardly be attainable... 시제?
the real-time decoder is hardly attainable with direct application of RNNLMs in place of traditional n-grams. 
%G 문장 분리.. 일부 수정..
In order to overcome such computational issues, most of the RNNLN systems adopt two-pass decoding strategy, which generates lattices or a set of n-best results based on n-gram in the first path, 
%G which가 여전히 주어...
and then performs the rescoring on the hypotheses with RNNLMs.

%Limitation of previous attempts: 1-pass cache - done
%G investigate는 타동사 (for 삭제)
Prior studies have investigated the possibility of implementing real-time decoder with RNNLM~\cite{onepass2}.
%G which는 RNNLM? studies? -> 분리
The study reduced computational complexity of the RNNLMs by caching the conditional probabilities of the words and 
%G results of feed-forwardings?
the results of RNN computation and reusing the cached data. 
%G it이 지칭하는 건?
However, even though the computational load was minimized  
%G In manner of reducing?
by introducing cache strategy and reducing redundant computations, 
%G it이 지칭하는 바가 없음..
%G away from -> far from
%G achieving decoder? the 삭제..
the result was still far from achieving real-time performance with large vocabulary based RNNLM.

%The strength of GPGPU in ASR
%G on the other hand... 어색
%G the wide range of ASR... 역시 어색.. the 삭제 및 area/field로 변경..
Recent studies have applied the General Purpose Graphic Processing Units (GPGPUs) in various fields of ASR~\cite{gpu3, gpu1, gpu2, gputraining}.
%G They는 누구?
One of the studies applied the GPGPU to training RNNLMs, and showed that the outstanding parallelization capability of GPGPU was suitable in minimizing the computational load of probability normalization processes~\cite{gputraining}.

%Main Problem
%G possibility of RNNLM?
In this paper, we investigate the possibility of implementing a GPGPU-based real-time Large Vocabulary Continuous Speech Recognition (LVCSR) that utilizes RNNLM. 
%G Get remedy of computation load... what it requires...
%In order to get remedy of computation load of it, we introduce the use of GPGPUs and what it requires under the many core framework.
%G ability -> capability
Even though GPGPUs have powerful parallelization capabilities, obstacles such as their insufficient memory size and slow data transfer speed between GPGPUs and CPUs 
%G RNN-based real-time decoder에 쓰는게... 흠...
discourage the use of GPGPUs among RNNLM-based real-time decoders.
%G the relatively에서 the 생략..
%G CPU에서 무엇을 어떻게 계산하는지가 전혀 없음...
%G 이 앞의 내용만 보면 마치 RNNLM 계산 자체를 GPU에서 하는 것으로 보이기도...
%Moreover, relatively slower speed of computations on CPUs could make whole processes slow down since GPGPUs have to wait until the computations on CPUs are finished even if GPGPUs have done their works.
Moreover, it is also important to balance the computation time between GPGPU and CPU, as the acceleration on GPGPU may not have a prominent impact on the overall speed if the GPGPU needs to wait for the CPU computation to finish.

%Main Idea
%G GPGPU의 단점을 해결한다는 표현은 좀 아닌거 같습니다... 특별히 문제가 있는 것도 아니니까... 차라리 optimize라면 모를까...
%G 사실상 RNNLM 자체에 GPGPU를 적용하는 것도 아니고... 그러고보면 앞의 RNNLM Training에 GPGPU를 적용한 예를 언급할 이유도 없었을지도...
%We try to apply GPGPUs to RNNLM-based network search by solving the disadvantages of GPGPUs.
In order to achieve real-time decoding of RNNLM-based LVCSR, we apply on-the-fly rescoring of RNNLM to GPGPU based network traversal technique proposed in ~\cite{gpu2}.
%G The goal이라는 것이 모호하여 아예 생략하였습니다.
We accelerate the speed of data exchange between the two heterogeneous processors and reduce redundant computations on CPUs by applying cache strategies.
%G a real-time speed -> real-time speed
The resulting recognition system has shown almost twice faster than real-time speed 
%G 그냥 in various circumstances라고 하니까 많은 경우에 그랬다 (일부 아니다)라는 느낌이 드는것도 같아서...
when experimented under various conditions, while maintaining relatively 10\% lower Word Error Rate (WER) than that of conventional n-gram models.

% Papaer oraganization
This paper is organized as the following.
In section 2, the structure of RNNLMs is explained. 
Section 3 explains how we applied GPGPUs to RNNLM-based network search.
Section 4 explains the RNNLM rescoring with caches.
Section 5 evaluates the improvement of the proposed method, followed by the conclusion in Section 6.
%REVIEW 3-2) 그림 1 수정
  	  \begin{figure}[t]
        \vspace{-17mm}
        \centering
        \includegraphics[bb=0 0 494 414,width=106mm,height=73mm]{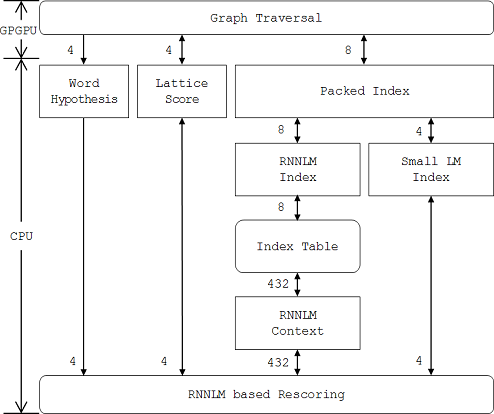}
        \label{fig:schematic_diagram_of_rnnlm}
        \caption{{\it The process of RNNLM-based graph traversals}}
      \end{figure}
  %Method
  \section{Recurrent Neural Network}
    In order to speed up RNNLM computations, we apply 
%G the->an (이미 알고있는 게 아니니...), strategies -> strategy
    an efficient RNNLM architecture described in this section, which consists of the hierarchically decomposed output layer~\cite{hs}, and Maximum Entropy (MaxEnt) strategy~\cite{me}.

    \subsection{Hierarchical Softmax}
    In RNNLM-based ASR system, directly computing the conditional probability of an input word 
%G an input word니까 a word sequence / complexity
for a given word sequence has high computational complexity 
%G it -> the system
since the system needs to normalize probabilities over the all the words in the vocabulary. 
%G In order to 와 which can 의 내용이 같아서 중복됩니다.. 둘 중의 하나만.
In order to alleviate the computational burden, the hierarchical softmax method was 
%G Introduce를 우리가 한게 아니니 apply 정도로 했습니다.
applied for our RNNLM implementation~\cite{hs2}.

	The output layer consists of a binary tree. We used Huffman tree method to build up the binary tree because it assigns shorter codes to more frequently used words 
%G fast가 수식하는게 training/decoding이 되는게 약간 어색해서 별도의 명사 speed를 붙였습니다.
and it leads to fast training and decoding speed~\cite{hs3}.
%G 그냥 normalization 대신에 Softmax를 넣었습니다.
%G probability 대신에 likelihood로 했습니다.
	In the output layer, the softmax normalization for computing the likelihood is performed only over the nodes on the path of the binary tree 
%G binary tree의 유일한 root를 지시한다는 의미로 보고 a root에서 the root로 변경
from the root to the input word node so that the computational cost is reduced 
%G scale이라는 말이 조금 애매해서 그냥 Big-O notation으로 변경하였습니다.
%from linear scale to logarithmic scale.
from $O(n)$ to $O(\log n)$, where $n$ represents the vocabulary size.
     
      \subsection{Maximum Entropy}
    In order to further reduce the computational cost, 
we interpolated hash-based MaxEnt models with the RNNLM itself.
We have used n-gram based MaxEnt models, which has similar accuracy to traditional n-gram maximum likelihood model and is easy to integrate with neural network configuration~\cite{me2}.
%G 다음 부분들이 전반적으로 불필요해 보이네요...
%    we apply hash-based MaxEnt models.
%    We use n-gram features to compute MaxEnts. 
%G have pp 로 갈 이유가 없어보임. just 역시 큰 의미 없음. as 대신에 similar to..
%G 그나저나 다음 문장 자체가 좀... 불필요한 사족 같기는 하네요.
%MaxEnt models with n-gram features show similar accuracy to traditional n-gram models~\cite{me2}.

Because the RNN-based and MaxEnt-based Language Models (LMs) are complementary to each other~\cite{me}, interpolating both LM scores enables us to reduce the size of the number of nodes in RNNLM without loss of accuracy.
%    We interpolate computation results of MaxEnts and main networks since the two models are complementary to each other~\cite{me} 
%G allows 대신에 enables가 나을 듯 합니다.
%so that it enables RNNLMs to have smaller hidden layers without loss of accuracy.
    Because the size of a hidden layer is a dominant factor for the computational complexity of RNNLMs,
%G we로 시작하는게 너무 많은거 같아서 변경...
the amount of computation can be reduced.
      
	\section{GPGPU Acceleration in RNNLM-based Graph Search}
	GPGPUs have been successfully applied in various fields of ASR by virtue of their characteristics of the powerful parallelism.
% to use -> in using
	However, there are some obstacles in utilizing GPGPUs for accelerating RNNLM-based network search.
% First, added
	First, GPGPUs have insufficient memory to load the whole content of large vocabulary RNNLMs. 
	%REVIEW 3-4) 영어표현
	Moreover, the data transfers between GPGPUs and CPUs are very time consuming works.
	This section explains how we apply GPGPUs to RNNLM-based graph traversals.

%TO-DO
	\subsection{On-the-fly RNNLM Rescoring}
We employ the GPGPU-CPU hybrid architecture as noted in~\cite{gpu2}.
%REVIEW 2-1) HCLG networks 들의 composition 임을 명시
%REVIEW 2-8) 2-1 에서 언급한 WFST 관련 설명을 추가한 것으로 조치, optimization 자체 설명은 미조치
%REVIEW 3-1) 2-1 에서 언급한 WFST 관련 설명을 추가한 것으로 조치
The RNNLM-based rescoring is deployed to both GPGPUs and CPUs in such a way that the Weighted Finite State Transducer (WFST) which is composition of HCLG networks. The G network is composed with a short span n-gram, which can be resided in the available GPGPU memory.
%REVIEW 2-2) 아래문장에서 설명이 된 것으로 보임 -> 미조치
While the frame-synchronous Viterbi search is performed to generate lattices on the GPGPU side, the LM portion of the lattice score is simultaneously rescored with RNNLM in on-the-fly manner on the CPU side.

    %Slow data transfer speed
%REVIEW 3-9) 공식화된 설명으로 변경 필요 -> 미조치
    \subsection{RNNLM Context Transfer}
% 수정...
Whenever a new word hypothesis is output from the WFST graph traversal on the GPGPU, the word hypothesis and its prior RNNLM context are sent to the CPU side so that the RNNLM computation can be done on the CPU. After the RNNLM computation, 
%REVIEW 3-5) 영어 표현 미조치
the resulting rescored score and the updated RNNLM context are sent back to GPGPU. 

%	Whenever the on-the-fly rescoring is performed, the word hypotheses and the RNNLM contexts which are needed for calculating RNNLMs, are sent to CPUs and after RNNLM computations, the results are sent back to GPGPUs. 
The exact size of each RNNLM context may depend on the structure of the RNNLM, but it is generally larger than a few hundreds of bytes. 
%	Moreover, the number of RNNLM look-ups can be larger by millions per utterance depending on the length of an input speech.
Moreover, the number of new word hypotheses per each frame can be as high as a few thousands, and so millions of RNNLM lookups may be requested per an utterance.
%REVIEW 3-6) 문의..
%REVIEW 3-7) 조금 약한 표현으로 변경 
	Considering the size of RNNLM context and the number of data exchange, the data transfer between GPGPU and CPU can cause speed degradation of RNNLM-based on-the-fly rescoring WFST traversals.
%	We overcame the problem not by exchanging the RNNLM contexts but by exchanging their indices.
We reduced the size of the data transfer by storing the RNNLM contexts on the CPU side, and only transferring the indices of the stored context to the GPGPU. 

	Figure~\ref{fig:schematic_diagram_of_rnnlm} depicts the proposed RNNLM-based graph traversal processes.
	Each number and arrow represents bytes and the flow of data, respectively.
%	When it comes to the size of RNNLM contexts, it will be explained in section 5.
	We have created an {\it IndexTable} for storing and retrieving RNNLM contexts and put the table into the CPU memory.
	The {\it IndexTable} is in charge of converting a large-sized RNNLM context into an 8-byte index, and vice versa.
	Because both encoding and decoding of the contexts need to be performed in a short time,
	we have made {\it IndexTable} bidirectional to handle both purposes.
%REVIEW 3-8) small LM 인덱스는 small lm의 ngram 을 지칭 한다고 설명	
	In addition to the RNNLM index, the index of the small LM, which corresponds to an ngram of the LM used to build the WFST graph, also needs to be transferred in order to compute and replace the small LM score with the RNNLM score.
	Instead of exchanging the two index sequences separately, we concatenate the two indices into one numerical value in order to further reduce the data size per each transaction.
%	The two indices are separately saved into the higher and lower order bits of the variable.	
%	We use 64 bits unsigned variables and assign 32 bits per each LM index so that the variable can cover a wide range of LM indices even if the number of RNNLM contexts gets larger by millions.
%REVIEW 2-3) "Compared to transmitting the whole context information including RNN hidden layer" 라는 것을 명시.
	With the indexing and packing method, we could reduce the exchanged data size to approximately $1/30$ compared to transmitting the whole context information including a recurrent layer.

     \section{RNNLM Rescoring with Cache}
In GPGPU-CPU hybrid architectures, the balance between the GPGPU and CPU speed is important, because relatively slower speed of CPUs can degrade the overall speed of RNNLM-based graph traversals since GPGPUs have to wait until the computations on CPUs are finished in order to work synchronously. Therefore, a fast RNNLM computation strategy on the CPU side is crucial in accelerating the overall on-the-fly rescoring time.
This section explains the efficient method to speed up RNNLM-based graph search on the hybrid architecture.

% Cache 구조 설명 : 
%    //Mapping Table 2 : Rnn Index
%    struct rnn_index{
%        int             word_idx;
%        unsigned int    state_idx;
%    };
%    struct rnn_result{
%        unsigned int    state_idx;
%        real            prob;
%    };
\subsection{The structure of RNNLM Context Cache}
%We have solved the problem by reducing the number of redundant computations on CPUs. 
We have optimized the computation on the CPU side by reducing the number of redundant computations during RNNLM calculation. 
%REVIEW 2-4) 영어표현
We use a cache strategy which stores the once-computed results and reuses them for the same input contexts.

Each element in the cache consists of a key-value pair: the key consists of a prior RNNLM context and a new word hypothesis, and the value consists of the resulting RNNLM score for the given word and the updated RNNLM context.
The RNNLM context generally consists of the values in the previous hidden layer, but we also use the previous word sequence for computing the MaxEnt portion of the RNNLM.
Since we are compressing the hundred bytes of context data into a 8-byte index by using {\it IndexTable} as explained in Section 3, we store the compressed indices instead of the whole context data.
%An input word index and a RNNLM context are needed to compute a RNNLM score.
%The RNNLM context consists of a previous hidden layer and a word sequence.
%Since the context data can be accessed with their indices by using {\it IndexTable} explained in section 3,
%we just store the indices of the context data.
%We have defined a structure which consists of an input word index and a RNNLM context index, 
%and used the structure as a key of the cache element.
%A result data of the RNNLM computation includes 
%the word probability and the resulting RNNLM context 
%which consists of a updated hidden layer and an input word index added word sequence.
%We have defined a structure which consists of the result data and used the structure as a value of the cache element.

%REVIEW 2-5) 3.2 3 번째 문단의 3번째 문장 부분에서 설명 된 것으로 보임 -> 미조치
%REVIEW 3-10) 필요이상으로 상세한 설명 -> 미조치
\subsection{RNNLM Probability Computation with Cache}
      \begin{table}[t]
        \vspace{2mm}
        \centerline{
          \begin{tabular}{|rl|}
            \hline
            \multicolumn{2}{|c|}{RnnlmProb($w$, $c$)} \\
            \hline \hline
               1 & $\mathbb{I}$ $\leftarrow$ ($w$, $c$) \\
               2 & if Cache[$\mathbb{I}$] exists then \\
               3 & \ \ \ \ $\mathbb{O}$ $\leftarrow$ Cache[$\mathbb{I}$] \\
               4 & else  \\
               5 & \ \ \ \ $\mathbb{C}$ $\leftarrow$ IndexTable[$c$] \\
               6 & \ \ \ \ ($p$,$\mathbb{C}^{\prime}$) $\leftarrow$ ComputeRnnlm($w$,$\mathbb{C}$) \\
               7 & \ \ \ \ if IndexTable$^{-1}$[$\mathbb{C}^{\prime}$] exists then \\
               8 & \ \ \ \ \ \ \ \ $c^{\prime}$ $\leftarrow$ IndexTable$^{-1}$[$\mathbb{C}^{\prime}$] \\
               9 & \ \ \ \ else \\
              10 & \ \ \ \ \ \ \ \ $c^{\prime}$ $\leftarrow$ lengthOf(IndexTable) + 1 \\
              11 & \ \ \ \ \ \ \ \ IndexTable[$c^{\prime}$] $\leftarrow$ $\mathbb{C}^{\prime}$ \\
              12 & \ \ \ \ end if \\
              13 & \ \ \ \ $\mathbb{O}$ $\leftarrow$ ($p$,$c^{\prime}$) \\
              14 & \ \ \ \ Cache[$\mathbb{I}$] $\leftarrow$ $\mathbb{O}$ \\
              15 & end if \\              
              16 & return $\mathbb{O}$ \\
            \hline
          \end{tabular}
        }
        \caption{\label{tab:cache_algorithm} {\it The process of RNNLM computation with caches}}
      \end{table}

% 테이블 기호 설명
Table~\ref{tab:cache_algorithm} shows a procedure for the probability computation of a word hypothesis in the proposed RNNLM-based network search.
$\mathbb{I}$ represents the key structure of the {\it Cache} element and it consists of the current RNNLM context index $c$ and the following word index $w$.
$\mathbb{O}$ represents the value structure of the {\it Cache} element and it consists of the LM probability $p$ and the updated RNNLM context index $c^{\prime}$.
%$\mathbb{C}^{\prime}$ represents the updated RNNLM context data.
{\it IndexTable} compresses the RNNLM context data $\mathbb{C}$ into an index variable $c$, and {\it Cache} stores the already computed pairs of inputs and outputs ($\mathbb{I}$, $\mathbb{O}$).

% 테이블 작동 설명
%The input parameters $w$ and $c$ are provided by the GPGPU network decoder explained in section 3.
%the frame-synchronous Viterbi search on GPGPUs.
The context index $c$ is associated with each path on the graph traversal on GPGPUs, and at every time the WFST network outputs a new word index $w$, the parameters $w$ and $c$ are sent back to the CPU side and fed into the RNNLM likelihood computation process.
At line 1 of Table~\ref{tab:cache_algorithm}, the two input parameters are stored to the input structure $\mathbb{I}$.
In lines 2--3, if $\mathbb{I}$ is already cached, then the retrieved value of {\it Cache}[$\mathbb{I}$] is saved to $\mathbb{O}$ and is returned without having to do any further computations.
Otherwise, in lines 4--6, the $\mathbb{C}$ is retrieved from the {\it IndexTable} with the index $c$, and the function {\it ComputeRnnlm} computes the conditional probability of $w$ based on the prior context $\mathbb{C}$, outputting the RNNLM probability value $p$ and the updated context $\mathbb{C}^{\prime}$. 

On a side note, the hidden layer value is initialized to a zero vector at the beginning of each utterance.
%REVIEW 2-6) 
The length of the word sequence in the RNNLM context $\mathbb{C}$ is restricted to the order of MaxEnt model, and the latest word sequence is maintained by removing the oldest word from the sequence whenever the number of previous words in the context exceeds the predefined size.

In lines 7--12, the updated context $\mathbb{C}^{\prime}$ is stored into the {\it IndexTable} and its index value $c^{\prime}$ is retrieved. When the context has been already stored before, the corresponding index is retrieved without adding it to the table again. 
%In lines 7-8, if {\it IndexTable} has $\mathbb{C}^{\prime}$, then $c^{\prime}$ is set to the retrieved index.
%Otherwise, in lines 9-12, $c^{\prime}$ is set to the 1 added length of {\it IndexTable} and $\mathbb{C}^{\prime}$ is saved to {\it IndexTable}[$c^{\prime}$]. 
In lines 13--15, $p$ and $c^{\prime}$ are saved to $\mathbb{O}$, and $\mathbb{I}$ and $\mathbb{O}$ are cached as a pair of key and value for later use.
%REVIEW 2-7) higher score 에서 lower score 를 뺀 값을 GPU로 보낸 다는 것 을 명시
Finally at line 16, $\mathbb{O}$ is returned. 
Difference between the returned value and the score of short span n-gram will be sent back to the GPGPU side and will be used to rescore the partially decoded WFST lattice.

	\section{Result}
	\subsection{Experimental Setup}
	%REVIEW 3-3) 미조치
	The experiments were performed both with the Wall Street Journal (WSJ) database and with a much larger set of data collected within the company ({\it in-house}).

	%AM
	The acoustic model for evaluating the in-house data was trained on 2,000 hours of the fully-transcribed Korean speech data.
	All the speech data were sampled at 16 kHz and were coded with 40-dimensional mel-frequency filterbank features, plus an additional dimension for the log-energy. The frames were computed every 10ms, and was windowed by 25ms Hamming window.
	Five frames to the left and right of the given frame were concatenated to the features to make a 451-dimensional acoustic feature vector in total.
	All the acoustic models were trained with the Deep Neural Network (DNN) which consisted of 5 hidden layers with 2,000 modes each. Rectified linear unit (ReLU) was used for the activation functions. The total number of output states was approximately 6,000.

	%LM
	The LM trained for evaluation of the in-house data was based on a total of 4GB text corpus, which amounts to approximately 74 million sentences with 475 million words, and the vocabulary size was about 1 million.
	The RNNLMs consist of one hidden layer with one hundred nodes, and the order of MaxEnt LM features was 3. 
	As for the n-gram models used for the comparison to the RNNLM, the 3-gram back-off models with Kneser-Ney smoothing were used.
	
	%WFST
	The WFST was compiled with a bigram LM and all the epsilon transitions were removed from the graph so that the computation on the GPGPU side can be optimized.
	%RNNLM to WFST 를 쓰지 않은 이유를 적어야하나?
	
	The evaluation tasks were performed on Intel Xeon X5690 3.47GHz processors with a total of 12 physical CPU cores and one Nvidia Tesla M2075 GPU equipped with 6GB memory.

\subsection{Speed up by Cache}
% Experiment 1: Cache Hit Ratio / Lookup Counts
% hit ratio 가 높은 이유는 조금 더 매끄러운 설명이 필요함.
By considering that the number of rescoring request per each utterance is as large as millions,
it was expected that naively applying RNNLM-based on-the-fly rescoring to ASR system would lead to high computational complexity.

As a matter of fact, our experiment showed that the naive application of the RNNLM to the on-the-fly rescoring decoder resulted in the speed slower than 10xRT. However, the RTF was significantly dropped to much lower than 1xRT when the cache strategy explained in section 4 was adopted. The hit ratio of the cache was at least 88.66\% over all the test cases, showing that most of the RNNLM computations were highly redundant.
%More precisely the average number of RNNLM computations was 233 thousands whereas that of looking up RNNLMs was 2.6 millions. 
%The reason why the cache hit ratios are high is that the activated word hypothesis list at a certain time are confined, the hypotheses are more likely to be activated.

	\subsection{Performance Evaluation}
      \begin{figure}[t]
        \vspace{-10mm}
        \centering
        \includegraphics[bb=0 0 433 275, width=106mm,height=60mm]{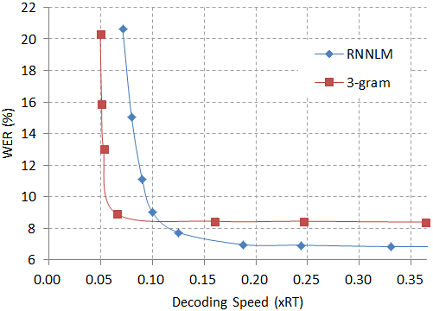}
        \caption{{\it Comparison of decoding speed and WER for RNNLM and 3-gram (in-house)}}
        \label{fig:eval92beam}
      \end{figure}

%Figure~\ref{fig:eval92beam} depicts relationship between the WER and decoding speed of both a RNNLM and a 3-gram model in in-house test case.
%In order to control decoding speed, we modified the beam size of the WFST traversal steps.
%In this experiment, the accuracy of both models reached their best at the decoding speed of 0.2--0.3xRT, and the WER of the RNNLM was relatively 10\% lower than that of the 3-gram model.
%Even if we set the beam size of the 3-gram model large enough, the WER of 3-gram did not get lower than 8.4\% whereas that of RNNLMs kept 6.8\%.
Figure~\ref{fig:eval92beam} depicts the performance comparison between RNNLM and 3-gram in the in-house 1 million vocabulary evaluation set. Various beam widths were applied to each LM in order to investigate how the word error rate changes with respect to the decoding speed.
The figure shows that while the lowest WER that 3-gram model can achieve is at around 8.49\%, RNNLM can reach as low as 6.83\% WER at the speed of 0.33 xRT.

%outperforms 3-gram in that the error rate of 3-gram is saturated to 8.49\% WER around approximately 0.05 xRT.
%On the contrary, that of RNNLMs increases beyond the speed where the performance of 3-gram is saturated.
%Eventually, we can achieve 6.83\% WER at 0.33 xRT.

       \begin{table}[t]
        \vspace{2mm}
        \centerline{
          \begin{tabular}{|c||cc|cc|cc|}
            \hline
			Type & \multicolumn{2}{c|}{in-house} & \multicolumn{2}{c|}{eval92} & \multicolumn{2}{c|}{dev93} \\
			Pass / Model      & \multicolumn{1}{c}{WER}& \multicolumn{1}{c|}{xRT}& \multicolumn{1}{c}{WER}& \multicolumn{1}{c|}{xRT}  & \multicolumn{1}{c}{WER}& \multicolumn{1}{c|}{xRT} \\
              \hline \hline
			\multicolumn{1}{|l||}{1 / 3-gram}  & 8.49 & 0.19 & 5.74 & 0.16 & 11.73 & 0.20 \\
			\multicolumn{1}{|l||}{1 / RNNLM}   & 6.83 & 0.64 & 4.15 & 0.43 & 10.80 & 0.40 \\
			\multicolumn{1}{|l||}{2 / hybrid}  & 7.57 & 0.33 & 5.74 & 0.25 & 11.31 & 0.27 \\
			\multicolumn{1}{|l||}{2 / RNNLM}  & 8.34 & 0.33 & 5.81 & 0.25 & 11.46 & 0.27 \\
            \hline
          \end{tabular}
        }
       \caption{\label{tab:multipletypesofrnnlm} {\it Performance comparison of different types of LM rescorings}}
      \end{table}

%Experiment 5: Comparison of Accuracy [N-Gram / RNN 2-pass / RNN Rescoring]
%REVIEW 1-1) 같은 디코딩 옵션을 사용했다고만 간단하게 설명.. 굳이 0.33 -> 0.64 까지 느리게 했는데 WER이 같다고 설명하지는 않음.
%REVIEW 2-9) knee of the curve 는 아니며, 다양한 한국어 테스트 셋으로 평가하여 선정한 옵션 값 BEAM 10 / Num-Hypothesis 10 을 사용하였음.
Table~\ref{tab:multipletypesofrnnlm} shows the performance comparison of four different rescoring methods with same decoding options which show the best performance all over the types of decodings. The one-pass types were computed by the proposed on-the-fly rescoring method, and the two-pass types were evaluated by rescoring the 1000-best hypotheses extracted from the lattices with 3-gram models. The two-pass hybrid type used interpolated score between 3-gram and RNNLMs.

%REVIEW 1-2) 문장 삭제
%REVIEW 2-10) 문장 삭제
%Slight but consistent improvement was observed in terms of recognition accuracy when we applied the proposed RNNLM-based WFST traversals.
Although the speed of the 1-pass RNNLM type was generally slower than that of other types, it was well within a real-time speed. This improvement shows that applying more accurate LMs at the WFST traversal stage at the first-pass leads to a better overall recognition result, instead of depending on the rescoring on the less accurate hypotheses generated from n-gram models.
%We expected that this improvement comes from the fact that more accurate LM was applied directly at the WFST traversal stage, instead of depending on the n-best hypotheses generated from n-gram models.

	\subsection{Memory Efficiency}

      \begin{table}[t]
        \vspace{2mm}
        \centerline{
          \begin{tabular}{| c@{}lrrr | r |}
            \hline
			& Total Memory & \# of Elem. & \multicolumn{1}{c|}{Size of Elem.} \\
            \hline \hline
			Hidden layer   & \multicolumn{1}{r}{400}    & \multicolumn{1}{r}{100}        & \multicolumn{1}{r|}{4} \\
			Word sequence  & \multicolumn{1}{r}{24}     & \multicolumn{1}{r}{3}          & \multicolumn{1}{r|}{8} \\
			MaxEnt Order   & \multicolumn{1}{r}{4}      & \multicolumn{1}{r}{1}          & \multicolumn{1}{r|}{4} \\
			RNNLM Index    & \multicolumn{1}{r}{4}      & \multicolumn{1}{r}{1}          & \multicolumn{1}{r|}{4} \\
            \hline
          \end{tabular}
        }
        \caption{\label{tab:cache_element} {\it The composition and the size of an element in IndexTable (in bytes)}}
      \end{table}

%Experiment 2: Cache Size per Utterance
Another measure to consider is the memory footprint of the {\it IndexTable}. Whereas {\it Cache} elements consists of only four integer values, each element in {\it IndexTable} is as large as a few hundred bytes and may increase rapidly as the decoding goes on.
In our experimental settings, each {\it IndexTable} item consists of four types of information and is as big as 432 bytes, as shown in Table~\ref{tab:cache_element}. 
%REVIEW 2-11) 영어표현
The size of the whole table increases during the decoding process and it is proportional to the number of unique elements.
%Another benefit of the high hit ratio was the efficient main memory usage. 
%In our experiment settings, each cache element consisted of four types of information listed in Table~\ref{tab:cache_element}. 
%Each element is saved twice for accessing by both RNNLM context indices and RNNLM context data as explained in section 4. The memory usage can be estimated as follows:
%      %
%      \begin{equation}
%      	E \times N \times 2
%      \end{equation}
%      %
%      where $E$ is the size of a cache element and $N$ is the number of caches. 
%In our evaluation tasks, the size of one element was 432 bytes and the average memory usage for caching per utterance was 191.78MB. More than 80\% of utterances over the all evaluation data sets were shorter than 10 seconds and the average cache size for the majority speeches was 169.23MB.

During the in-house evaluation task, the average memory usage for {\it IndexTable} per utterance was 191.78MB. More than 80\% of utterances over all evaluation data sets were shorter than 10 seconds, and average utterance length was about 7.69 sec. Considering that the size of the {\it IndexTable} will be generally proportional to the length of the utterance, we can reasonably assume that the memory footprint for the {\it IndexTable} is contained within acceptable size. %and the average cache size for the majority speeches was 169.23MB.

	\subsection{Caching Over Multiple Utterances}

      \begin{table}[t]
        \vspace{2mm}
        \centerline{
          \begin{tabular}{| r@{}llll | r |}
            \hline
			\multicolumn{1}{|c}{Capacity(KB)} & \ \ \ \# of Cache&\multicolumn{1}{c}{Mem.(MB)}&\multicolumn{1}{c|}{xRT}	\\
            \hline \hline
			\multicolumn{1}{|r}{0(=per utt)} & \multicolumn{1}{r}{69K}&\multicolumn{1}{r}{57.30}&\multicolumn{1}{r|}{0.64}\\
			\multicolumn{1}{|r}{250}&\multicolumn{1}{r}{164K}&\multicolumn{1}{r}{135.43}& \multicolumn{1}{r|}{0.59} \\
			\multicolumn{1}{|r}{500}&\multicolumn{1}{r}{282K}&\multicolumn{1}{r}{232.86}& \multicolumn{1}{r|}{0.57} \\
			\multicolumn{1}{|r}{750}&\multicolumn{1}{r}{406K}&\multicolumn{1}{r}{334.70}& \multicolumn{1}{r|}{0.57} \\
			\multicolumn{1}{|r}{1000}&\multicolumn{1}{r}{560K}&\multicolumn{1}{r}{461.49}& \multicolumn{1}{r|}{0.57} \\
            \hline
          \end{tabular}
        }
      \caption{\label{tab:cache4multipleutterances} {\it Cache memory usage and decoding speed depending on the capacity of cache (in-house) }}
      \end{table}

%Experiment 3: Cache for Multiple Utterances
We noticed that many speakers tend to repeat similar commands in different utterances, and so we hypothesized that maintaining the cache and {\it IndexTable} over multiple utterances may increase the hit ratio of the cache and will decrease the decoding speed further. Therefore, instead of resetting the caches for each utterance, we set a certain size boundary for the number of cache to be maintained, and kept the information over multiple utterances. We expected that a larger cache size will lead to a faster decoding speed, due to increased cache hit ratio.
%We tried to maintain caches for multiple utterances. The evaluation task was performed on in-house test case.

Table~\ref{tab:cache4multipleutterances} shows the decoding speed for different cache capacity. The memory usages of caches and the number of caches are averaged over all the utterances in the test set.

%REVIEW 1-3) Cache 사이즈를 늘림에 따라 0.64 -> 0.57 빨라지는 것을 1.2 배 라고 잘못 표기 하여 1.12 배로 정정
However, as can be seen from the table, the decoding speed did not get much faster than 1.12x even though the capacity of cache increased. We concluded that the result reflects the fact that even if the users are prone to speak the similar commands repeatedly, most of the word hypotheses are different for different utterances, and so the cache hit ratio does not get higher. %Thus we decided to reset the cache for every utterance.
%maintaining the caches for multiple utterances
Although the multi-utterance cache strategy only showed a marginal improvement, it would still be meaningful to adopt a smart cache strategy that limits the size of the cache by removing the cached items that are least frequently used, so that the size of the cache table does not overflow.

  \section{Conclusions}

This paper explained how we applied RNNLMs to a real-time large vocabulary decoder by introducing the use of GPGPUs.
We tried to accelerate RNNLM-based WFST traversals in GPGPU-CPU hybrid architectures by solving some practical issues for applying GPGPUs.
Moreover, in order to minimize the computation burden on CPUs, we applied a cache strategy.
%REVIEW 1-4) Table 2 를 보면 상대적으로 10% 향상 되는 것을 알 수 있으므로 숫자까지 표기하지는 않음.
The decoding speed of RNNLMs was still slower than that of n-gram models, but the proposed method achieved the real-time speed while maintaining relatively 10\% lower WER as shown in Table~\ref{tab:multipletypesofrnnlm},
and so we could perfectly apply this approach to an on-line streaming speech recognition engine. 
The memory footprint for the cache method was small enough to perform the experiment on the large data set. However, it will be more desirable to employ efficient cache techniques to reduce the memory usage further.

\newpage
  \eightpt
  \bibliographystyle{IEEEtran}
  %\bibliography{mybib}

\end{document}